\pdfoutput=1

\documentclass[11pt]{article}

\usepackage{naacl2021}

\usepackage{times}
\usepackage{latexsym}
\usepackage{booktabs}
\usepackage{amsmath}
\usepackage{amsfonts}
\usepackage{graphicx}
\usepackage{balance}
\usepackage[T1]{fontenc}

\usepackage[utf8]{inputenc}

\usepackage{microtype}

\newcommand\bo{\mathbf{o}}

\newcommand\eg{\textit{e.g.}}
\newcommand\ie{\textit{i.e.}}

\title{BERT4SO: Neural Sentence Ordering by Fine-tuning BERT}

\author{Yutao Zhu$^{1}$, Jian-Yun Nie$^{1}$, Kun Zhou$^{2}$, Shengchao Liu$^{1,3}$, Yabo Ling$^{1}$, Pan Du$^{1}$ \\
  $^{1}$Université de Montréal, Montréal, Québec, Canada \\
  $^{2}$School of Information, Renmin University of China, Beijing, China \\
  $^{3}$Mila - Québec Artificial Intelligence Institute, Montréal, Canada \\
  \texttt{\{yutao.zhu,yabo.ling\}@umontreal.ca, \{nie,pandu\}@iro.umontreal.ca} \\
  \texttt{francis\_kun\_zhou@163.com, liusheng@mila.quebec} \\
  \\
}

\begin{document}
\maketitle
\begin{abstract}
Sentence ordering aims to arrange the sentences of a given text in the correct order. Recent work frames it as a ranking problem and applies deep neural networks to it. In this work, we propose a new method, named BERT4SO, by fine-tuning BERT for sentence ordering. We concatenate all sentences and compute their representations by using multiple special tokens and carefully designed segment (interval) embeddings. The tokens across multiple sentences can attend to each other which greatly enhances their interactions. We also propose a margin-based listwise ranking loss based on ListMLE to facilitate the optimization process. Experimental results on five benchmark datasets demonstrate the effectiveness of our proposed method.
\end{abstract}

\section{Introduction}
Sentence ordering is the task of arranging sentences into an order so as to maximize the coherence of the text~\cite{DBLP:journals/coling/BarzilayL08}. 
This task has been widely studied due to its significance in downstream tasks, such as determining the ordering of concepts in concept-to-text generation~\cite{DBLP:conf/acl/KonstasL12,DBLP:conf/emnlp/KonstasL13}, information from various document in extractive multi-document summarization~\cite{DBLP:journals/jair/BarzilayEM02,DBLP:conf/aaai/NallapatiZZ17}, and events in story generation~\cite{DBLP:conf/acl/FanLD19,DBLP:conf/acl/ZhuSDNZ20}. 

Early studies on sentence ordering generally use handcrafted linguistic features to model the document structure~\cite{DBLP:conf/acl/Lapata03,DBLP:conf/naacl/BarzilayL04,DBLP:journals/coling/BarzilayL08}, which limits the application of these systems. Recent works apply neural networks to model the text coherence and solve the sentence ordering task. Typical methods are based on pointer network~\cite{DBLP:conf/nips/VinyalsFJ15}, which leverages attention as a pointer to select successively a member of the input sequence as the output. These methods~\cite{DBLP:journals/corr/GongCQH16,DBLP:conf/emnlp/CuiLCZ18,DBLP:conf/aaai/Wang019a,DBLP:conf/ijcai/YinSSZZL19} usually require sentence-by-sentence decoding to produce the reordered sentences. One drawback of such approaches is that the current time step prediction depends on the previous predictions, making it difficult for ordering a large set of sentences.

\begin{table}[t!]
    \centering
    \small
    \begin{tabular}{p{7.3cm}}
    \toprule
        (2) When they arrived they saw some airplanes in the back of a truck. \\
        (3) The \textbf{kids} had a hard time deciding what to ride first. \\
        (1) The family got together to go to the fair. \\
        (5) Finally they played the dart game. \\
        (4) Then \textbf{they played} some games to win prizes. \\
    \bottomrule
    \end{tabular}
    \caption{An example of unordered sentences in a document and their correct order is on the left.}
    \vspace{-10pt}
    \label{tab:eg}
\end{table}

More recently, ranking-based frameworks and pre-trained language models have been applied to sentence ordering~\cite{DBLP:conf/aaai/KumarBKR20,DBLP:conf/acl/PrabhumoyeSB20}. Different from sequential prediction, the ranking-based framework aims at predicting a global ranking score for each sentence and computing the order by sorting the scores. 
Sentence ordering can also benefit from the pre-trained language models (\eg, BERT~\cite{DBLP:conf/naacl/DevlinCLT19}), which enhance sentence representations. 
\citet{DBLP:conf/acl/PrabhumoyeSB20} used BERT to judge the relative order between sentence pairs and applies a topological sort algorithm to infer the entire order. 
This method achieves the state-of-the-art performance, but with a high computational cost because it enumerates all sentence pairs.
\citet{DBLP:conf/aaai/KumarBKR20} proposed a method where each sentence is encoded by BERT and the interaction between sentences is captured by a transformer based on the sentence representations. The limitation here is that the separately encoded sentences cannot take the cross-sentence interactions between tokens into account. However, it is common that clues of sentence orders can be revealed by enabling across-sentence tokens to attend to each other.
Taking the fourth sentence in Table 1 for example, ``they'' usually contributes little information in an isolated sentence representation, but when connected with ``kids'' in the third sentence, it becomes a strong indication of the sentence orders. 


In this work, we propose a new structure to capture cross-sentence interactions between tokens for \textbf{S}entence \textbf{O}rdering by fine-tuning \textbf{BERT} (which is called \textbf{BERT4SO}), and design a new listwise objective function accordingly. Instead of encoding each sentence separately, we propose to concatenate all sentences as a long sequence and leverage multiple [CLS] tokens to represent the sentences. Every token from any sentence can attend to others and their interaction information can thus be captured. Extensive experiments on five benchmark datasets demonstrate the effectiveness of our design. Besides, to further facilitate the optimization of our method, we also propose a margin-based ListMLE which successfully improve our method on small datasets. 

\begin{figure}[t!]
    \centering
    \includegraphics[width=\linewidth]{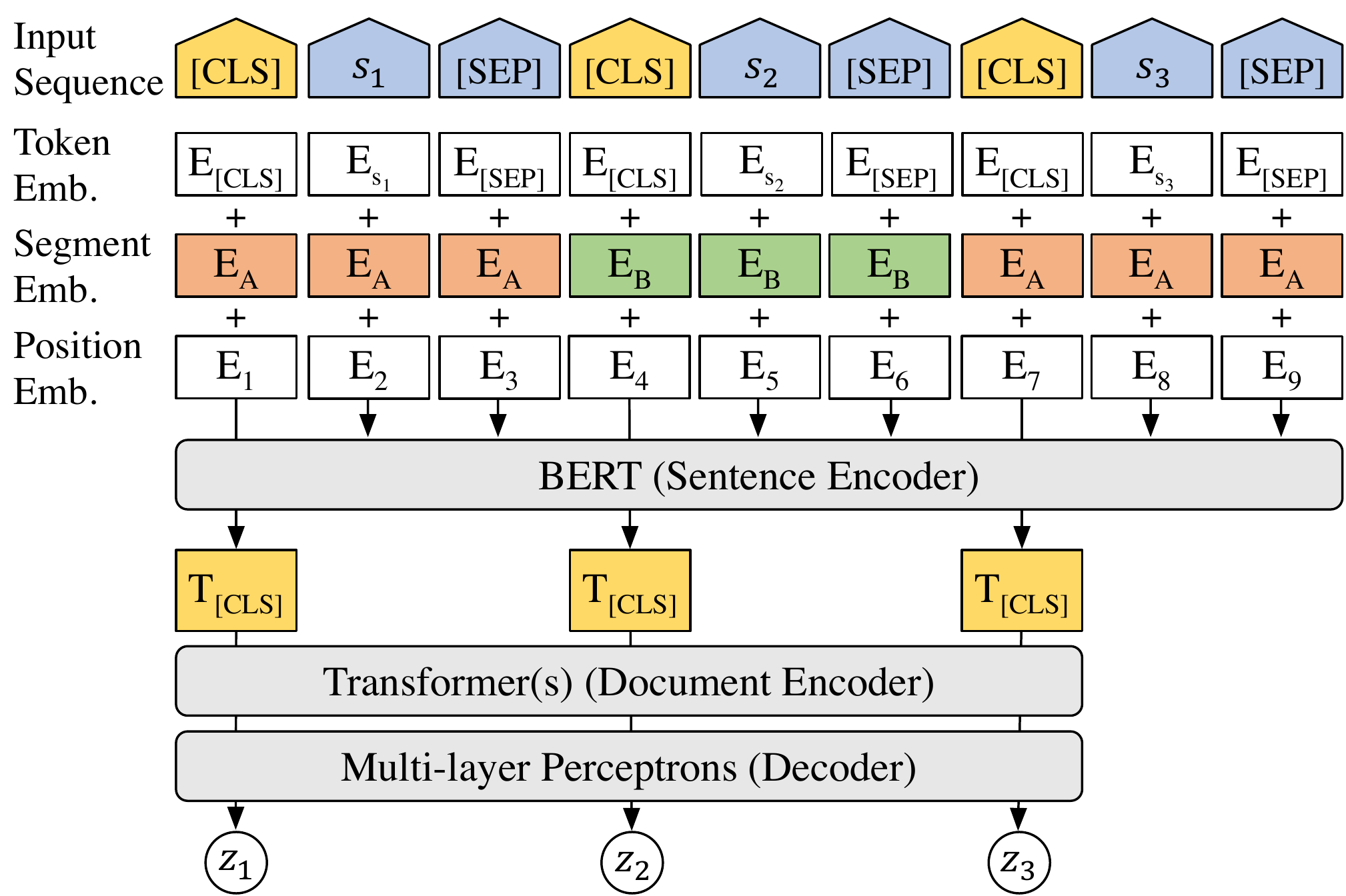}
    \caption{The structure of our method.}
    \label{fig:model}
    \vspace{-10pt}
\end{figure}

\section{Methodology}
Assume that we have a set of $n$ sentences $\{s_{o_1},\cdots,s_{o_{n}}\}$ with a random order $\bo=[o_1,\cdots,o_{n}]$, our task is to find the right order for these sentences $\bo^*=[o_1^*, \cdots, o_{n}^*]$.
Following the existing work~\cite{DBLP:conf/aaai/KumarBKR20,DBLP:conf/acl/PrabhumoyeSB20,DBLP:journals/corr/abs-2101-11178}, this task is framed as a ranking problem, where the model is trained to predict a score $z_i$ for each sentence $s_i$, and the global order is determined by sorting it.

\subsection{Fine-tuning BERT for Sentence Ordering}
BERT~\cite{DBLP:conf/naacl/DevlinCLT19} is a pre-trained language model trained on several large corpora.
Fine-tuning BERT has been successfully used in many downstream tasks such as question answering~\cite{DBLP:conf/emnlp/KhashabiMKSTCH20} and machine reading comprehension~\cite{DBLP:conf/acl/LiuHCG19,DBLP:conf/acl/ClarkLKML19}.
Since BERT is pre-trained on tasks with only one or two sentences, it cannot be directly applied to sentence ordering with multiple sentences. Existing work~\cite{DBLP:conf/aaai/KumarBKR20} alleviated this problem by encoding each sentence separately, but the interaction between tokens across sentences cannot be well-captured. In contrast, we propose to concatenate all sentences into a long sequence and use two segment embeddings in BERT to indicate their intervals. In so doing, each token can attend to all others so that the interaction among tokens can be captured. Moreover, after obtaining the contextual representations of sentences, we apply another multi-layer transformer to create the sentence representations at document level, which further capture the interactions among sentences. The final representation is used to calculate a score, and the global order is determined by sorting it.

\noindent\textbf{Multiple Sentences Encoding} As illustrated in Figure~\ref{fig:model}, we insert a [CLS] token before each sentence and a [SEP] token after each sentence, and then concatenate all sentences as a long one. 
In the original BERT model, the [CLS] token is used to aggregate the features from a sentence or a sentence pair.
We modify it by leveraging multiple [CLS] tokens to get representations for multiple sentences. 
Before feeding into BERT, three embeddings are used to represent a token: (1) Token embeddings are provided by pre-trained BERT. (2) In vanilla BERT, two segment embeddings ($\text{E}_A$ and $\text{E}_B$) are used to distinguish different sentences. To deal with multiple sentences within a document in our task, we propose to concatenate these segment embeddings alternatively to segment the whole sequence, \eg, for a document with three sentences $[s_1, s_2, s_3]$, the segment embeddings will be $[\text{E}_A, \text{E}_B, \text{E}_A]$. (3) Position embeddings are used to indicate the position of each token. 
The sum of the three embeddings is used as input to the BERT model. Then we use the output at the [CLS] position as the representation of each sentence. 

\noindent\textbf{Sentence Ordering}
After obtaining the sentence representations, we stack several transformer layers as \textit{document encoder} to further enhance the interaction between sentences and capture the document-level features. Different from sentence encoder, the input to document encoder is the sentence representations, thus it can capture more global perspective features.
Finally, for each sentence $s_i$, we compute the final score $z_i$ by a \textit{decoder} with two multi-layer perceptrons (MLPs) based on the sentence representations computed by the document encoder.

\subsection{Margin-based ListMLE}
ListMLE is a listwise ranking loss based on the relative order between each sentence and its following sentences\footnote{The computation of ListMLE is given in Appendix.}. It has been proven to be more effective~\cite{DBLP:conf/aaai/KumarBKR20} than pointwise or pairwise losses in optimizing the transformer-based sentence ordering methods.
Inspired by the idea of margin maximization in pairwise ranking~\cite{DBLP:conf/kdd/Joachims02}, we incorporate the idea into listwise loss in order to avoid underfitting and achieve a better convergence rate~\cite{DBLP:journals/pr/WangZWH18}. For a document with $n_i$ unordered sentences, assuming the correct order is $\bo_i^*=[o_1^*,\cdots,o_{n_i}^*]$, we propose the following modified margin-based listwise loss:
\begin{align}
F_j(k) & = \frac{\exp(z_{o_k^*})}{\sum_{l=j}^{n_i}\exp(z_{o_l^*})}, \label{eq:confidence}\\
\tilde f(j) &= \log F_j(j) + \sum_{k=j+1}^{n_i-1} \log (\gamma - F_j(k)), \label{eq:max_margin}\\
\mathcal{L}(\theta) &= -\sum_{i=1}^{N} \sum_{j=1}^{n_i-1} \frac{1}{n_i - j} \tilde f(j),
\end{align}
where $\gamma$ is a hyperparameter. Eq.~\ref{eq:confidence} is the normalized score of each sentence, and Eq.~\ref{eq:max_margin} aims at enlarging the margin of the correctly ordered sentence in $j$-th position, while lowering the margin of other sentences.

\section{Experiments}
\subsection{Datasets and Evaluation Metrics}
We conduct experiments on five datasets\footnote{The download links can be found in the original paper. The detailed statistics of datasets and implementation of our model are presented in Appendix.}. 

\noindent\textbf{NeurIPS/AAN/NSF abstracts}~\cite{DBLP:conf/aaai/LogeswaranLR18}. These datasets consist of abstracts from NeurIPS, ACL and NSF research award papers, including 3,259, 12,157, and 127,865 samples, respectively. The data are split into training, validation and test set according to the publication year. 

\noindent\textbf{SIND captions}~\cite{DBLP:conf/naacl/HuangFMMADGHKBZ16}. This is a visual story dataset with 50,200 stories. Each story contains five sentences. It is split into training, validation, and test set with the ratio of 8:1:1.

\noindent\textbf{ROCStory}~\cite{DBLP:journals/corr/MostafazadehCHP16}. It is a commonsense story dataset with 98,161 stories. Each story comprises five sentences. We make an 8:1:1 random split on the dataset to get the training, validation and test set.

We use Kendall's $\tau$ and Perfect Match Ratio (PMR) as the evaluation metrics, both are commonly used in previous work~\cite{DBLP:journals/corr/GongCQH16,DBLP:conf/aaai/LogeswaranLR18,DBLP:conf/aaai/KumarBKR20}.

\noindent\textbf{Kendall's Tau} ($\tau$): it is one of the most frequently used metrics for text coherence evaluation~\cite{DBLP:journals/coling/Lapata06, DBLP:conf/aaai/LogeswaranLR18}. 
It measures how much a ranking agrees with the ground-truth.

\noindent\textbf{PMR}: it calculates the percentage of samples for which the entire order of the sequence is correctly predicted~\cite{DBLP:journals/corr/ChenQH16}. 

\begin{table*}[t!]
    \centering
    \small
    \setlength{\tabcolsep}{0.15cm}{
    \begin{tabular}{lcccccccccc}
    \toprule
        & \multicolumn{2}{c}{NeurIPS} & \multicolumn{2}{c}{AAN} & \multicolumn{2}{c}{NSF} & \multicolumn{2}{c}{SIND} & \multicolumn{2}{c}{ROCStory} \\
        \cmidrule(lr){2-3} \cmidrule(lr){4-5} \cmidrule(lr){6-7} \cmidrule(lr){8-9} \cmidrule(lr){10-11} 
        & ${\tau}$ & PMR & ${\tau}$ & PMR & ${\tau}$ & PMR & ${\tau}$ & PMR & ${\tau}$ & PMR \\
    \midrule
        CNN + PtrNet$^\heartsuit$ & 0.6976$^\dag$ & 19.36$^\dag$ & 0.6700$^\dag$ & 28.75$^\dag$ & 0.4460$^\dag$ & 5.95$^\dag$ & 0.4197$^\dag$ & 9.50$^\dag$ & 0.6538$^\dag$ & 27.06$^\dag$ \\
        LSTM + PtrNet$^\heartsuit$ & 0.7373$^\dag$ & 20.95$^\dag$ & 0.7394$^\dag$ & 38.30$^\dag$ & 0.5460$^\dag$ & 10.68$^\dag$ & 0.4833$^\dag$ & 12.96$^\dag$ & 0.6787$^\dag$ & 28.24$^\dag$ \\
        Variant-LSTM + PtrNet$^\heartsuit$ & 0.7258$^\dag$ & 22.02$^\dag$ & 0.7521$^\dag$ & 40.67$^\dag$ & 0.5544$^\dag$ & 10.97$^\dag$ & 0.4878$^\dag$ & 13.57$^\dag$ & 0.6852$^\dag$ & 30.28$^\dag$ \\
        ATTOrderNet$^\heartsuit$ & 0.7466$^\dag$ & 21.22$^\dag$ & 0.7493$^\dag$ & 40.71$^\dag$ & 0.5494$^\dag$ & 10.48$^\dag$ & 0.4823$^\dag$ & 12.27$^\dag$ & 0.7011$^\dag$ & 34.32$^\dag$ \\
        HierarchicalATTNet$^\diamondsuit$ & 0.7008$^\dag$ & 19.63$^\dag$ & 0.6956$^\dag$ & 30.29$^\dag$ & 0.5073$^\dag$ & 8.12$^\dag$ & 0.4814$^\dag$ & 11.01$^\dag$ & 0.6873$^\dag$ & 31.73$^\dag$ \\
        SE-Graph$^\diamondsuit$ & 0.7370$^\dag$ & 24.63$^\ddag$  & 0.7616$^\dag$ & 41.63$^\dag$ & 0.5602$^\dag$ & 10.94$^\dag$ & 0.4804$^\dag$ & 12.58$^\dag$ & 0.6852$^\dag$ & 31.36$^\dag$ \\
        ATTOrderNet + TwoLoss$^\diamondsuit$ & 0.7357$^\dag$ & 23.63$^\ddag$ & 0.7531$^\dag$ & 41.59$^\dag$ & 0.4918$^\dag$ & 9.39$^\dag$ & 0.4952$^\dag$ & 14.09$^\dag$ & 0.7302$^\dag$ & 40.24$^\dag$ \\
        RankTxNet$^\heartsuit$ & 0.7684$^\dag$ & 26.12$^\ddag$ & 0.7744$^\dag$ & 38.84$^\dag$ & 0.4899$^\dag$ & 6.81$^\dag$ & 0.5528$^\dag$ & 14.80$^\dag$ & 0.7333$^\dag$ & 30.19$^\dag$\\
        B-TSort$^\diamondsuit$ & \textbf{0.7884} & 30.59 & 0.8064 & \textbf{48.08} & 0.4813$^\dag$ & 7.88$^\dag$ & 0.5632$^\dag$ & 17.35$^\ddag$ & 0.7941$^\dag$ & 48.06$^\dag$ \\
    \midrule
        BERT4SO & 0.7778 & \textbf{30.70} & \textbf{0.8076} & 45.41 & \textbf{0.6379} & \textbf{13.00} & 0.5916 & \textbf{19.07} & \textbf{0.8487} & \textbf{55.65} \\
        \quad with ListMLE & 0.7516 & 24.13 &  0.8045 & 44.42 & 0.6344 & 12.78 & \textbf{0.5998} & 18.83 & 0.8468 & 55.04 \\
    \bottomrule
    \end{tabular}
    }
    \caption{Results on five benchmark datasets.  Models with $\diamondsuit$ are implemented by provided source code while those with $\heartsuit$ are implemented by ourselves. The numbers here are our runs of the model. $\dag$ and $\ddag$ denote significant improvements with our method in t-test with $p < 0.01$ and $p < 0.05$ respectively.}
    \label{tab:re}
    \vspace{-10pt}
\end{table*}

\subsection{Baseline Models}
\textbf{Various Pointer Network based Methods}: CNN/LSTM+PtrNet~\cite{DBLP:journals/corr/GongCQH16}, Variant-LSTM+PtrNet~\cite{DBLP:conf/aaai/LogeswaranLR18}, ATTOrderNet~\cite{DBLP:conf/emnlp/CuiLCZ18}, HierarchicalATTNet~\cite{DBLP:conf/aaai/Wang019a}, SE-Graph~\cite{DBLP:conf/ijcai/YinSSZZL19}, and ATTOrderNet+TwoLoss~\cite{DBLP:conf/aaai/YinMSGSZL20} all adopt CNN/RNN to obtain the representation for the input sentences and employ the pointer network as the decoder to predict order.

\textbf{RankTxNet}~\cite{DBLP:conf/aaai/KumarBKR20} applies BERT and transformers to order the sentences. Different from ours, this model encodes each sentence separately and is trained by the ListMLE loss.

\textbf{B-TSort}~\cite{DBLP:conf/acl/PrabhumoyeSB20}\footnote{Note that the results of B-TSort are slightly worse than those reported in the original paper, because the provided source code does not shuffle the sentence order on test set which arbitrarily improves the results.} is the state-of-the-art model in sentence ordering. The model applies BERT to judge the relative order between each pair of sentences and build a graph based on the relative ordering. The global order is inferred by a topological sort algorithm on the graph. 




\subsection{Experimental Results}
Table~\ref{tab:re} shows the results of all models on the five datasets. We have the following observations:

(1) BERT4SO significantly outperforms all baselines on NSF, SIND, and ROCStory and achieves comparable results with the previous best method B-TSort on NeurIPS and AAN. Specifically, on ROCStory, BERT4SO outperforms B-TSort by around 5.5\% $\tau$ and 7.6\% PMR. This result clearly demonstrates the effectiveness and wide applicability of our proposed method.
(2) RankTxNet, B-TSort, and BERT4SO apply BERT for encoding sentences. We can see that they achieve better performance than other RNN-based ones. This demonstrates the superiority of applying BERT for sentence ordering.
Furthermore, BERT4SO achieves the best performance, indicating that it can better leverage BERT for sentence ordering than a naive utilisation. 
(3) Compared with RankTxNet, BERT4SO concatenates all sentences so that they can attend to each other in token-level. The higher performance confirms this design can better capture sentence relationship and the token-level interaction, which are beneficial to sentence ordering.
(4) B-TSort performs similarly to BERT4SO on NeurIPS and AAN, showing its effectiveness on small datasets. B-TSort enumerates all sentence pairs and estimates their relative order. It leverage exhaustive pairwise ordering information for the whole set, but the process is very expensive and difficult to apply when documents contain lots of sentence.
Indeed, the training time of B-TSort is extremely long (\eg, $11\times$ more time than ours on NSF and $7\times$ more time on SIND). 

\begin{table}[t!]
    \centering
    \small
    \begin{tabular}{lcccc}
        \toprule 
        & \multicolumn{2}{c}{Original} & \multicolumn{2}{c}{Ensemble} \\
        \cmidrule(lr){2-3} \cmidrule(lr){4-5}
        & $\tau$ & PMR & $\tau$ & PMR \\
        \midrule
        NeurIPS & 0.7778 & 30.70 & \textbf{0.7916} & \textbf{30.85} \\
        AAN & 0.8076 & 45.41 & \textbf{0.8265} & \textbf{48.62} \\
        NSF & \textbf{0.6379} & \textbf{13.00} &  0.5884 & 11.65 \\
        \bottomrule
    \end{tabular}
    \caption{Results of BERT4SO trained on ensemble datasets of NeurIPS, AAN and NSF.}
    \label{tab:ens}
    \vspace{-15pt}
\end{table}
\noindent\textbf{Discussion}
To evaluate the performance of our proposed margin-based ListMLE, we also train BERT4SO with the traditional listMLE. The results are shown in Table~\ref{tab:re}. We can observe that both loss functions work well on three large datasets (\ie, NSF, SIND, and ROCStory), but BERT4SO with the traditional ListMLE cannot achieve good results on the very small dataset NeurIPS. We speculate that the improvements stem from the second term in margin-based ListMLE (Eq.~\ref{eq:max_margin}), which adds more constraints on lowering the scores of other sentences except the current ground-truth, thus makes more effective use of each sample. 

We also find the improvements of BERT4SO are limited on small datasets. To alleviate this problem, inspired by a recent study~\cite{DBLP:conf/lrec/LiuZL20}, we combine the training set of the three similar datasets NeurIPS, AAN, and NSF and test the model on each test set, respectively.
The results in Table~\ref{tab:ens} clearly show that, training on the combined dataset brings improvements for both NeurIPS and AAN because NSF provides lots of additional training data. This result confirms that BERT4SO performs well when provided with sufficient training data. On the other hand, the performance on NSF drops when combined with NeurIPS and AAN data. We believe that in this case, the addition of other datasets to NSF does not add more useful training examples, but more noise. 

\section{Conclusion and Future Work}
In this work, we proposed a new method for sentence ordering by fine-tuning BERT and a modified ListMLE loss. Our proposed structure greatly enhanced the cross-sentence interactions, thus obtained improvements on five benchmark datasets. The newly proposed loss function is shown to be helpful on optimizing our method for small datasets. We also found that the performance on small datasets can be improved by combining similar datasets. In the future, we will investigate other methods, such as self-supervised learning, to improve our model on small datasets.

\clearpage
\balance

\clearpage
\appendix
\section{ListMLE}
ListMLE~\cite{DBLP:conf/icml/XiaLWZL08} is a surrogate loss to the perfect order 0-1 based loss function. Given a corpus with $N$ documents, the $i^\text{th}$ document with $n_i$ unordered sentences is denoted by $d_i=[s_1,\cdots,s_{n_i}]$. Assume the correct order of $d_i$ is $\bo_i^*=[o_1^*,\cdots,o_{n_i}^*]$, then ListMLE is computed as:
\begin{align}
    \mathcal{L}(\theta) &= -\sum_{i=1}^{N}\log f(\bo_i^*|d_i), \\
    f(\bo_i^*|d_i) &= \prod_{j=1}^{n_i}\frac{\exp(z_{o_j^*})}{\sum_{k=j}^{n_i}\exp(z_{o_k^*})}.
\end{align}
With ListMLE, the model will assign the highest score for the first sentence and the lowest score for the last one.

\begin{table}[t]
    \centering
    \small
    \setlength{\tabcolsep}{1.1mm}{
    \begin{tabular}{lrrrrrr}
    \toprule
        \textbf{Datasets} & \textbf{Max.} & \textbf{Avg.} & \textbf{Train} & \textbf{Val.} & \textbf{Test} \\
    \midrule
        NeurIPS ab. & 512 & 181.73 & 2,448 & 409 & 402 \\
        AAN ab. & 1,030 & 134.95 & 8,569 & 962 & 2,626 \\
        NSF ab. & 2,923 & 263.62 & 96,017 & 10,185 & 21,573 \\
        SIND captions & 288 & 58.61 & 40,155 & 4,990 & 5,055 \\
        ROCStory & 100 & 52.84 & 78,529 & 9,816 & 9,816 \\
    \bottomrule 
    \end{tabular}
    }
    \caption{The statistics of all datasets. Max. and Avg. stands for the maximum and average number of tokens in documents.}
    \label{tab:sta}
\end{table}
\section{Statistics of Datasets}
The statistics of all datasets are shown in Table~\ref{tab:sta}.

\section{Implement Details}
Our model is implemented by PyTorch~\cite{DBLP:conf/nips/PaszkeGMLBCKLGA19} and HuggingFace's Transformers~\cite{DBLP:journals/corr/abs-1910-03771}. We train it on a TITAN V GPU with 12GB memory. We test \{1,2,3\} Transformer layers for the document encoder and choose two layers due to its best performance on validation set. The hidden size of the decoder is 200. AdamW optimizer~\cite{DBLP:conf/iclr/LoshchilovH19} is applied for training. The learning rate is 5e-5 for sentence encoder (BERT) and document encoder (multi-layer transformers), and 5e-3 for MLPs. The batch size is 32. The model is trained with 5 epochs and 20\% of all training steps are used for learning rate warming up. The margin hyperparameter $\gamma$ is tuned in \{0.25, 0.5, 0.75, 1\} and set as 1. All models are selected according to their performance (\ie, the sum of $\tau$ score and PMR score) on validation set.

\end{document}